\documentclass[lettersize,journal]{IEEEtran}

\usepackage{amssymb}
\usepackage{amssymb}
\usepackage{graphicx}      

\usepackage{amsthm}

\usepackage[tbtags]{amsmath}

\usepackage{epstopdf}
\usepackage{subfigure}
\usepackage{comment}
\usepackage{multirow}
\usepackage{ctable}
\usepackage[flushleft]{threeparttable}
\usepackage{array}
\usepackage[linesnumbered,lined,boxed,algoruled]{algorithm2e}
\SetKwInOut{Input}{input}
\SetKwInOut{Output}{output}
\SetKwComment{Comment}{/* }{ */}
\usepackage{setspace}
\usepackage{diagbox}

\usepackage{amsmath,amsfonts}
\usepackage{array}
\usepackage[caption=false,font=normalsize,labelfont=sf,textfont=sf]{subfig}
\usepackage{textcomp}
\usepackage{stfloats}
\usepackage{url}
\usepackage{verbatim}
\usepackage{graphicx}
\usepackage{cite}
\begin{document}

\title{Improving the Robustness of Deep Convolutional Neural Networks Through Feature Learning}

\author{{Jin Ding, Jie-Chao Zhao, Yong-Zhi Sun, Ping Tan, Ji-En Ma, You-Tong Fang}
\thanks{This work was supported by the National Natural Science Foundation of China (No.51977193). (\emph{The corresponding author: Jin Ding})}
\thanks{J. Ding, J. Zhao, Y. Sun, and P. Tan are with School of Automation and Electrical Engineering \& Key Institute of Robotics of Zhejiang Province, Zhejiang University of Science and Technology, Hangzhou, 310023, China (e-mail: jding@zust.edu.cn; zjiechao@126.com; sunyongzhi@hotmail.com; tanp@supcon.com)}
\thanks{J. Ma and Y. Fang are with School of Electrical Engineering, Zhejiang University, Hangzhou, 310027, China (e-mail: majien@zju.edu.cn; youtong@zju.edu.cn)}}



\maketitle

\begin{abstract}
Deep convolutional neural network (DCNN for short) models are vulnerable to examples with small perturbations. Adversarial training (AT for short) is a widely used approach to enhance the robustness of DCNN models by data augmentation. In AT, the DCNN models are trained with clean examples and adversarial examples (AE for short) which are generated using a specific attack method, aiming to gain ability to defend themselves when facing the unseen AEs. However, in practice, the trained DCNN models are often fooled by the AEs generated by the novel attack methods. This naturally raises a question: can a DCNN model learn certain features which are insensitive to small perturbations, and further defend itself no matter what attack methods are presented. To answer this question, this paper makes a beginning effort by proposing a shallow binary feature module (SBFM for short), which can be integrated into any popular backbone. The SBFM includes two types of layers, i.e., Sobel layer and threshold layer. In Sobel layer, there are four parallel feature maps which represent horizontal, vertical, and diagonal edge features, respectively. And in threshold layer, it turns the edge features learnt by Sobel layer to the binary features, which then are feeded into the fully connected layers for classification with the features learnt by the backbone. We integrate SBFM into VGG16 and ResNet34, respectively, and conduct experiments on multiple datasets. Experimental results demonstrate, under FGSM attack with $\epsilon=8/255$, the SBFM integrated models can achieve averagely 35\% higher accuracy than the original ones, and in CIFAR-10 and TinyImageNet datasets, the SBFM integrated models can achieve averagely 75\% classification accuracy. The work in this paper shows it is promising to enhance the robustness of DCNN models through feature learning.

\end{abstract}


\section{Introduction}
It is well known that deep convolutional neural network (DCNN for short) models can be fooled by examples with small perturbations \cite{Goodfellow2015ICLR, Szegedy2014ICLR}, which results in serious consequences when they are applied into the safety-critical applications, e.g., autonomous driving, airport security, and industry automation. Therefore, it is of great significance to enhance the robustness of DCNN models.

Adversarial training (AT for short) is a widely used approach to enhance the robustness of DCNN models by data augmentation \cite{Andriushchenko2020NIPS, Bai2021IJCAI, Carlini2017SP, Wang2022ICML}. In each training step, one specific attack method is employed to generate adversarial examples (AE for short), which together with the clean examples are input into DCNN models, expecting to make the trained DCNN models defend themselves on unseen AEs. However, the trained models can still be fooled by AEs figured out by the novel attack methods. This naturally raises a question: can a DCNN model learn certain features which are insensitive to small perturbations, and further defend itself no matter what attack methods are presented.

\begin{figure}[h!]
\centering
\subfigure[Clean cat]{\includegraphics[width=0.2\textwidth]{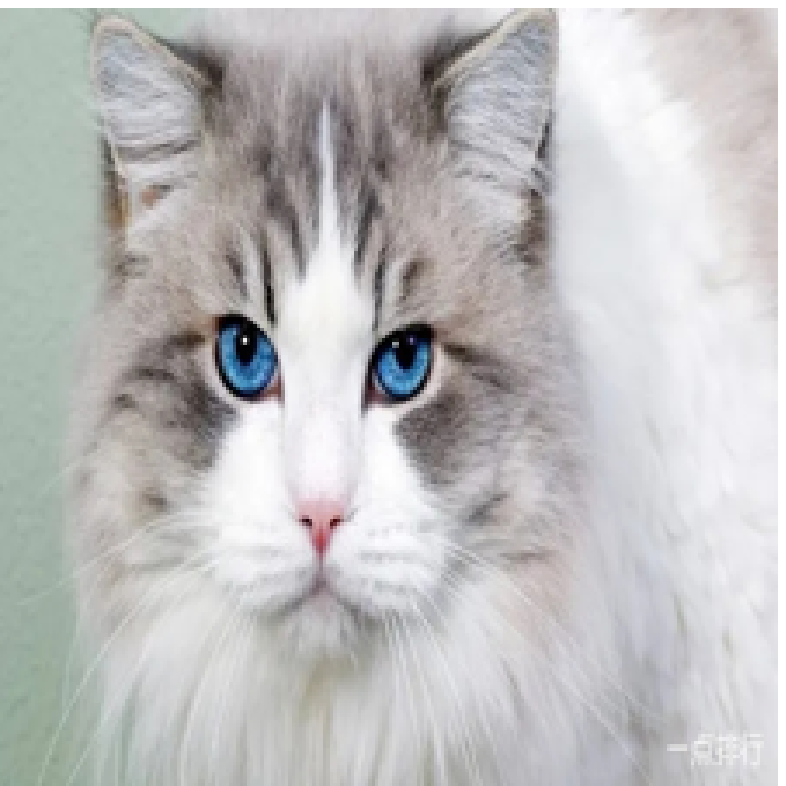}\label{fig:ORI}}
\hspace{0.05\textwidth}
\subfigure[Adversarial cat]{\includegraphics[width=0.2\textwidth]{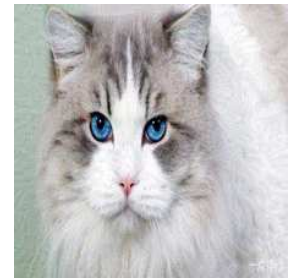}\label{fig:AE}}
\caption{\textbf{Clean example and adversarial example}.}
\label{fig:ORIAE}
\end{figure}

Fig.~\ref{fig:ORIAE} shows a clean example and its adversarial counterpart. With the presence of small noise, the DCNN models amplify the noise through the deep structures, which is reduced into the final texture features, and probably output the wrong classification results. However, comparing Fig.~\ref{fig:ORI} and Fig.~\ref{fig:AE}, it is clear to see the shape of the cat is legible in both images, which prompts us the shape-like features may be helpful in making a correct decision with the presence of the small noise. In this regard, Li \emph{et al.} \cite{Li2023TPAMI} proposed a part-based recognition model with human prior knowledge, which first segments the objects, scores the object parts, and outputs the classification class based on the scores. Sitawarin \emph{et al.} \cite{Sitawarin2022ArX} also proposed a part-based classification model, which combines part segmentation with a classifier. Both works take parts of objects into consideration to enhance adversarial robustness of DCNN models. The results are outstanding, but they require the detailed segmentation annotations and cannot be integrated into popular recognition or detection architectures.

\begin{figure}[h!]
\centering
\subfigure[Thresholded edge image for clean cat]{\includegraphics[width=0.2\textwidth]{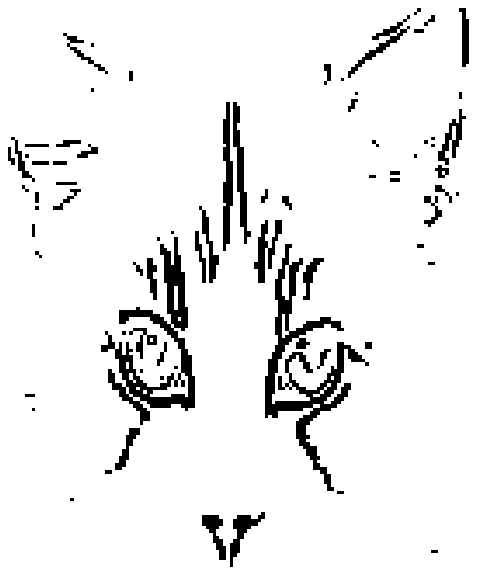}\label{fig:ORIBI}}
\hspace{0.05\textwidth}
\subfigure[Thresholded edge image for adversarial cat]{\includegraphics[width=0.2\textwidth]{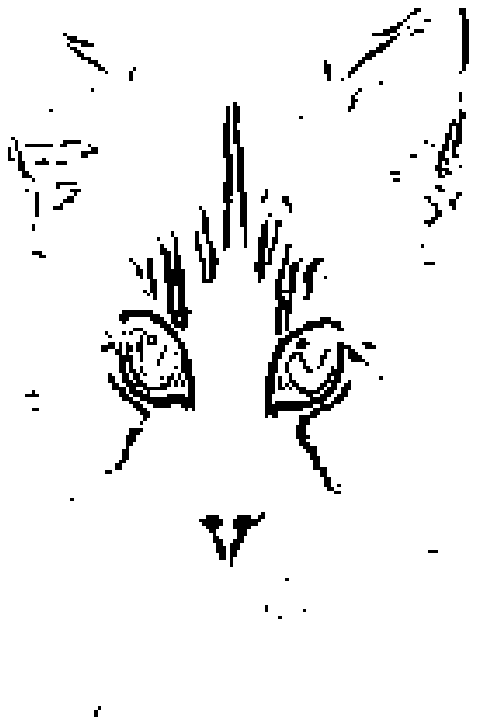}\label{fig:AEBI}}
\caption{\textbf{Thresholded edge images of Fig.~\ref{fig:ORIAE}}.}
\label{fig:ORIAEBI}
\end{figure}

Fig.~\ref{fig:ORIAEBI} shows the thresholded edge images of Fig.~\ref{fig:ORIAE}. We can see both binary images are almost the same, which indicates binary features can be taken as shape-like features to enhance the robustness of DCNN models. In this paper, a shallow binary feature module (SBFM for short) is proposed to extract the binary features of the input images, which stacks two types of layers-\--Sobel layer and threshold layer. In Sobel layer, there are four parallel feature maps which represent horizontal, vertical, and diagonal edge features respectively \cite{Kittler1983IVC, Xiao2022JOS}. And in threshold layer, it turns the learnt edge features from Sobel layers to the binary features, which then are feeded into the fully connected layers for classification with the features learnt by the backbones.
SBFM is lightweight and can be integrated into any popular backbone, e.g., VGG \cite{Simonyan2014ICLR} and ResNet \cite{He2016CVPR}.
We integrate SBFM into VGG16 and ResNet34, respectively. The experimental results on CIFAR-10, TinyImageNet, and Cats-and-dogs datasets show the training accuracy, test accuracy, and training time of the SBFM integrated models are on a par with those of original ones, and when attacked by FGSM \cite{Goodfellow2015ICLR} with $\epsilon=8/255$, the SBFM integrated models can achieve averagely 35\% higher classification accuracy than the original ones, and 75\% classification accuracy in CIFAR-10 and TinyImageNet datasets. The results demonstrate the binary features learnt by SBFM can enhance the robustness of DCNN models notably.

The contributions of this paper can be summarized as follows:
\begin{enumerate}
  \item We first raise the question "can a DCNN model learn certain features which are insensitive to small perturbations, and further defend itself no matter what attack methods are presented", and make a beginning effort on this.
  \item SBFM is proposed to learn binary features through Sobel layer and threshold layer, which is lightweight and can be integrated into any popular backbone.
  \item Experimental results on multiple datasets show, when attacked by FGSM with $\epsilon=8/255$, the SBFM integrated models can achieve averagely 35\% higher accuracy compared to the original ones, and 75\% classification accuracy n CIFAR-10 and TinyImageNet datasets.
\end{enumerate}


\section{Related Work}
\textbf{Adversarial Training.}
AT is a widely used approach to enhance the robustness of DCNN models by generating AEs during each training step.
Tsipras \emph{et al.} \cite{Tsipras2019ICLR} found there exists a tradeoff between robustness and standard accuracy of the models generated by AT, due to the robust classifiers learning a different feature representation compared to the clean classifiers.
Madry \emph{et al.} \cite{Madry2018ICLR} proposed an approximated solution framework to the optimization problems of AT, and found PGD-based AT can produce models defending themselves against the first-order adversaries.
Kannan \emph{et al.} \cite{Kannan2018NIPS} investigated the effectiveness of AT at the scale of ImageNet \cite{Deng2009CVPR}, and proposed a logit pairing AT training method to tackle the tradeoff between robust accuracy and clean accuracy.
Wong \emph{et al.} \cite{Wong2020ICLR} accelerated the training process using FGSM attack with random initialization instead of PGD attack \cite{Kurakin2017ICLR}, and reached significantly lower cost.
Xu \emph{et al.} \cite{Xu2022NIPS} proposed a novel attack method which can make a stronger perturbation to the input images, resulting in the robustness of models by AT using this attack method is improved.
Li \emph{et al.} \cite{Li2022CVPR} revealed a link between fast growing gradient of examples and catastrophic overfitting during robust training, and proposed a subspace AT method to mitigate the overfitting and increase the robustness.
Dabouei \emph{et al.} \cite{Dabouei2022ECCV} found the gradient norm in AT is higher than natural training, which hinders the training convergence of outer optimization of AT. And they proposed a gradient regularization method to improve the performance of AT.
AT improves the robustness of the DCNN models by generating AEs using one specific attack method. However, in practice, the models trained by AT may be still vulnerable when facing the novel attack methods.

\textbf{Part-based Model.}
Li \emph{eg al.} \cite{Li2023TPAMI} argued one reason that DCNN models are prone to be attacked is they are trained only on category labels, not on the part-based knowledge as humans do. They proposed a object recognition model, which first segments the parts of objects, scores the segmentations based on human knowledge, and final outputs the classification results based on the scores.
This part-based model shows better robustness than classic recognition models across various attack settings.
Sitawarin \emph{et al.} \cite{Sitawarin2022ArX} also thought richer annotation information can help learn more robust features. They proposed a part segmentation model with a head classifier trained end-to-end. The model first segments objects into parts, and then makes predictions based on the parts.

In both works, the shape-like features are leant to enhance the robustness of DCNN models remarkably. However, the proposed part-based models require the detailed segmentation annotation of objects and prior knowledge, and are hard to be combined with the existing recognition or detection architectures. In this paper, we propose a binary feature learning module which can be integrated into any popular backbone and enhance the robustness of DCNN models notably.

\section{Shallow Binary Feature Module}
\begin{figure}[h!]
    \centering
    \includegraphics[width=0.4\textwidth]{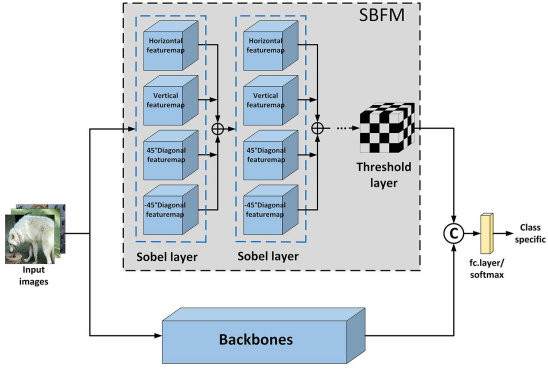}
    \caption{\textbf{The architecture of a SBFM integrated classification model}.}
    \label{fig:Archi}
\end{figure}

In this section, we give a detailed description of SBFM, which is lightweight and can be integrated into any popular backbone. SBFM stacks two types of layers-\--Sobel layer and threshold layer. In Sobel layer, four parallel feature maps are learnt to represent horizontal, vertical, and diagonal features respectively. In threshold layer, it sets a threshold to turn the features from Sobel layers into binary features.
The architecture of a SBFM integrated classification model is shown in Fig.~\ref{fig:Archi}.

\begin{figure}[h!]
\centering
\subfigure[Horizontal kernel]{\includegraphics[width=0.2\textwidth]{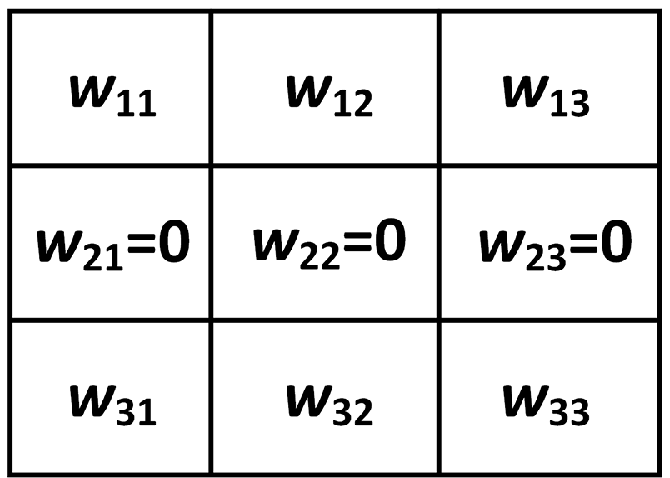}\label{fig:Khori}}
\hspace{0.05\textwidth}
\subfigure[Vertical kernel]{\includegraphics[width=0.2\textwidth]{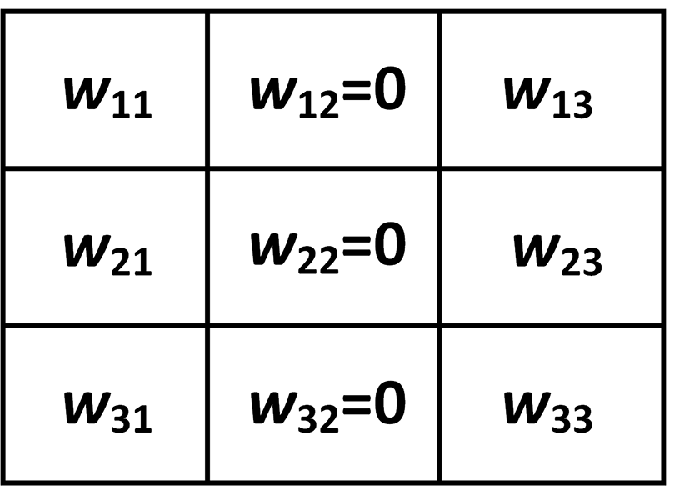}\label{fig:Kvert}}
\subfigure[Positive diagonal kernel]{\includegraphics[width=0.2\textwidth]{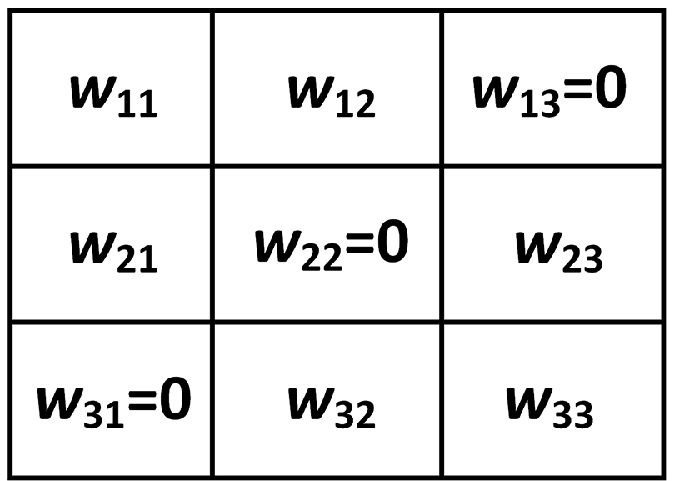}\label{fig:Kpodi}}
\hspace{0.05\textwidth}
\subfigure[Negative diagonal kernel]{\includegraphics[width=0.2\textwidth]{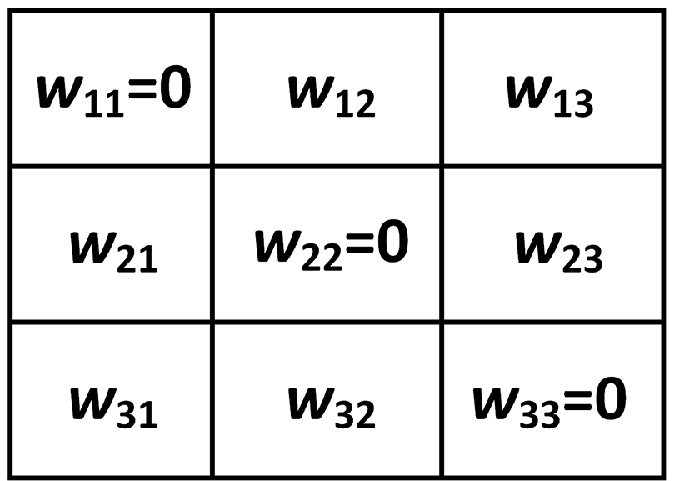}\label{fig:Knedi}}
\caption{\textbf{Four types of kernels of Sobel layer}.}
\label{fig:Kernel}
\end{figure}

\textbf{Sobel Layer. }
A Sobel layer consists of four parallel feature maps, i.e., horizontal feature maps, vertical feature maps, positive diagonal feature maps, and negative diagonal feature maps. For each feature map, a constrained kernel is designed to ensure the corresponding features can be learnt. Taking kernel size of 3 by 3 for example, the four types of kernel are shown in Fig.~\ref{fig:Kernel}.

For horizontal kernel in Fig.~\ref{fig:Khori},
\begin{equation}
\label{eq:Khori}
w_{ij}\left\{
\begin{array}{ll}
\in \textrm{[0, 1]} & \mbox{if \emph{i} = 1, \emph{j} = 1, 2, 3},\\
= \textrm{0} & \mbox{if \emph{i} = 2, \emph{j} = 1, 2, 3},\\
\in \textrm{[-1, 0]} & \mbox{if \emph{i} = 3, \emph{j} = 1, 2, 3},\\
\end{array}
\right.
\end{equation}

For vertical kernel in Fig.~\ref{fig:Kvert},
\begin{equation}
\label{eq:Kvert}
w_{ij}\left\{
\begin{array}{ll}
\in \textrm{[0, 1]} & \mbox{if \emph{j} = 1, \emph{i} = 1, 2, 3},\\
= \textrm{0} & \mbox{if \emph{j} = 2, \emph{i} = 1, 2, 3},\\
\in \textrm{[-1, 0]} & \mbox{if \emph{j} = 3, \emph{i} = 1, 2, 3},\\
\end{array}
\right.
\end{equation}

For positive diagonal kernel in Fig.~\ref{fig:Kpodi},
\begin{equation}
\label{eq:Kpodi}
w_{ij}\left\{
\begin{array}{ll}
\in \textrm{[0, 1]} & \mbox{if (\emph{i}, \emph{j}) $\in$ \{(1, 1), (1, 2), (2, 1) \}},\\
= \textrm{0} & \mbox{if (\emph{i}, \emph{j}) $\in$ \{(1, 3), (2, 2), (3, 1) \}},\\
\in \textrm{[-1, 0]} & \mbox{if (\emph{i}, \emph{j}) $\in$ \{(2, 3), (3, 2), (3, 3) \}},\\
\end{array}
\right.
\end{equation}

For negative diagonal kernel in Fig.~\ref{fig:Knedi},
\begin{equation}
\label{eq:Knedi}
w_{ij}\left\{
\begin{array}{ll}
\in \textrm{[0, 1]} & \mbox{if (\emph{i}, \emph{j}) $\in$ \{(1, 2), (1, 3), (2, 3) \}},\\
= \textrm{0} & \mbox{if (\emph{i}, \emph{j}) $\in$ \{(1, 1), (2, 2), (3, 3) \}},\\
\in \textrm{[-1, 0]} & \mbox{if (\emph{i}, \emph{j}) $\in$ \{(2, 1), (3, 1), (3, 2) \}},\\
\end{array}
\right.
\end{equation}

The output of a Sobel layer is the addition of the four feature maps.

\textbf{Threshold Layer. }
Features computed from the last Sobel layer are feeded into threshold layer, which can be denoted as $\textbf{x}^{in}$ $\in$ $R^{N\times P\times Q}$, where $N$ is the number of channels of the feature map, and $P$ and $Q$ are the width and height of each channel respectively.
We set a threshold for each channel, and the features output from the threshold layer, $\textbf{x}^{out}$, can be computed using Eq.~\ref{eq:featureoutput}.

\begin{equation}
\label{eq:featureoutput}
x_{ikj}^{out}=\left\{
\begin{array}{ll}
1 & \mbox{if $x_{ikj}^{in}$ $\geq$ $t$ * $\max$($\textbf{x}_{i}^{in}$)},\\
0 & \mbox{if $x_{ikj}^{in}$ $<$ $t$ * $\max$($\textbf{x}_{i}^{in}$)},\\
\end{array}\\
\right.
\end{equation}
where $i$ = 1,2,...,$N$, $k$ = 1,2,...,$P$, and $j$ = 1,2,...,$Q$. $t$ $\in$ [0, 1], and $\max$($\textbf{x}_{i}^{in}$) represents the maximum value of channel $i$.

It is obvious to see that, the higher $t$, the less binary features obtained; the smaller $t$, the more binary features obtained.

\section{Experiments}\label{sec:Expe}
In this section, we conduct experiments to examine whether the SBFM integrated VGG16 (VGG16-SBFM) and ResNet34 (ResNet34-SBFM) are more robust than the original ones on CIFAR-10 \cite{Krizhevsky2009}, TinyImageNet \cite{Le2015Tiny}, and Cats-and-Dogs \cite{Elson2007ACM} datasets.
Firstly, the effects of SBFM on model training are described. Secondly, the robustness of the SBFM integrated models and the original ones are compared under FGSM attack. And thirdly, the impacts of the number of Sobel layers and threshold on the robustness of the SBFM integrated models are analyzed.
All experiments are coded using TensorFlow, and run on Intel i9 cpu of 3.6GHz with 64 GB RAM. One TITAN XP GPU is employed.

\subsection{Datasets}
\textbf{CIFAR-10. }CIFAR-10 dataset includes 10 classes, each of which has 6,000 training images, and 1,000 test images. In experiments, 10\% training images in each class are taken as validation images. The resolution of images is 32$\times$ 32.

\textbf{TinyImageNet. }TinyImageNet dataset includes 200 classes, each of which has 500 training images, and 50 validation images. In experiments, validation images are taken as test images, and 10\% training images in each class are taken as validation images. The resolution of images is 64$\times$ 64.

\textbf{Cats-and-Dogs. }Cats-and-Dogs dataset includes 2 classes, each of which has 12,500 images. In experiments, 9,000, 1,000, and 2,500 images in each class are taken as training, validation, and test images, respectively. The resolution of images is resized to 224$\times$ 224.

\subsection{Effects of SBFM on model training}\label{sec:EXTraining}
We examine the effects of SBFM on model training by comparing three metrics between the SBFM integrated models and the original ones, i.e., training accuracy, test accuracy, and training time per epoch.
In SBFM, there are generally two parameters to be set-\--one is the number of Sobel layers, denoted as $l$, and the other is the proportion coefficient for threshold layer, denoted as $t$.
On CIFAR-10 dataset, $l$ and $t$ are set to 3 and 0.8, respectively, for both VGG16-SBFM and ResNet34-SBFM.
On TinyImageNet and Cats-and-Dogs dataset, $l$ and $t$ are set to 2 and 0.8 for VGG16-SBFM, and 3 and 0.8 for ResNet34-SBFM, respectively.
The loss function, optimizer, the batch size, and the number of epochs are set to be the same.
Each model is run for five times, and the average values are recorded.
Table~\ref{tab:effectsSBFM} shows the comparison of training performance between SBFM integrated models and the original ones.
From the table, it is clear to see the SBFM integrated models are on a par with the original models on three metrics, which indicates SBFM has no side effect on model training and is lightweight.

\begin{table*}[ht]
\begin{center}
\begin{threeparttable}

\begin{tabular}{l|ccc|ccc|ccc}
\hline
\multirow{2}*{} & \multicolumn{3}{|c|}{\textbf{CIFAR-10}} & \multicolumn{3}{|c|}{\textbf{TinyImageNet}} & \multicolumn{3}{|c}{\textbf{Cats-and-Dogs}}  \\
		\cline{2-10}
		~ & \textbf{Tr.Acc.} & \textbf{Te.Acc.} & \textbf{TimePE} & \textbf{Tr.Acc.} & \textbf{Te.Acc.} & \textbf{TimePE} & \textbf{Tr.Acc.} & \textbf{Te.Acc.} & \textbf{TimePE} \\
\hline
\textbf{VGG16} & 99.17\% & 83.70\% & 18s & 94.32\% & 33.84\% & 80s & 97.64\% & 80.10\% & 14s \\
\textbf{VGG16-SBFM} & 99.53\% & 82.95\% & 18s & 92.03\% & 30.01\% & 95s & 97.05\% & 83.17\% & 23s \\
\hline
\textbf{ResNet34} & 99.32\% & 79.19\% & 48s & 97.53\% & 30.80\% & 238s & 96.32\% & 78.30\% & 85s \\
\textbf{ResNet34-SBFM} & 99.48\% & 74.58\% & 48s & 97.98\% & 26.31\% & 250s & 95.64\% & 73.80\% & 92s \\
\hline
\end{tabular}
\begin{tablenotes}
\item \textbf{Tr.Acc.} stands for training accuracy. \textbf{Te.Acc.} stands for test accuracy. \textbf{TimePE} stands for training time per epoch.
\end{tablenotes}

\end{threeparttable}
\end{center}

\caption{Comparison of training performance between SBFM integrated models and the original ones.}
\label{tab:effectsSBFM}
\end{table*}

\subsection{Performance comparison between SBFM integrated models and original models under FGSM attack}
We compare the classification accuracy between SBFM integrated models and original ones under FGSM attack, with $\epsilon$ set to 0.1/255, 0.5/255, 1/255, 2/255, 3/255, 5/255, and 8/255.
$l$ and $t$ of the SBFM integrated models are set to the same values in Sec.~\ref{sec:EXTraining}.
Each model is run for five times, and the average value is recorded.

\begin{figure}[h!]
    \centering
    \includegraphics[width=0.5\textwidth]{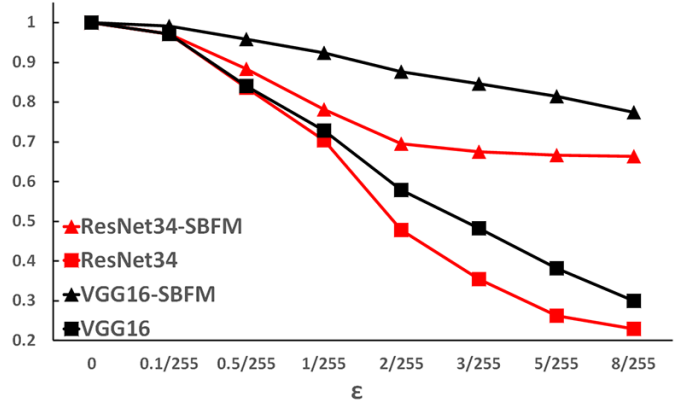}
    \caption{\textbf{Changing curve of classification accuracy with the strengthening of the attack intensity $\epsilon$ on CIFAR-10}.}
    \label{fig:FGSMCIFAR}
\end{figure}

\begin{figure}[h!]
    \centering
    \includegraphics[width=0.5\textwidth]{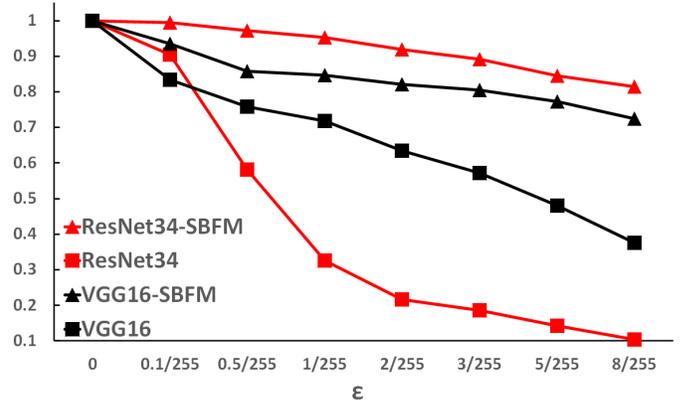}
    \caption{\textbf{Changing curve of classification accuracy with the strengthening of the attack intensity $\epsilon$ on TinyImageNet}.}
    \label{fig:FGSMTINY}
\end{figure}

\begin{figure}[h!]
    \centering
    \includegraphics[width=0.5\textwidth]{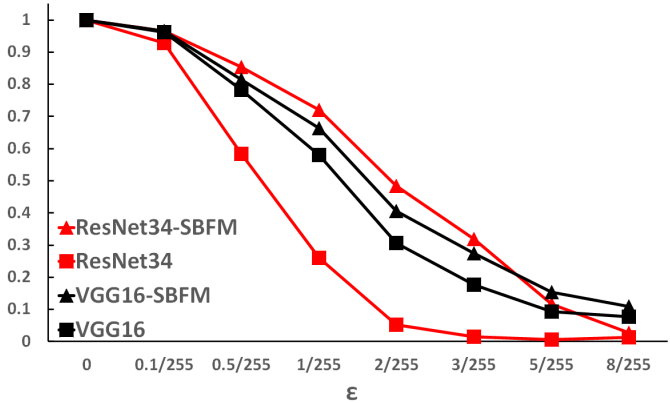}
    \caption{\textbf{Changing curve of classification accuracy with the strengthening of the attack intensity $\epsilon$ on Cats-and-Dogs}.}
    \label{fig:FGSMCATS}
\end{figure}

Fig.~\ref{fig:FGSMCIFAR}-Fig.~\ref{fig:FGSMCATS} show the changing curves of the classification accuracy for both SBFM integrated models and the original ones with the attack intensity $\epsilon$ increasing. From the figures, we can see that, in both CIFAR-10 and TinyImageNet datasets, the changing curves of SBFM integrated models are much more gentle compared to the original ones with the increase of $\epsilon$, showing the robustness is enhanced greatly after integrating SBFM. In Cats-and-Dogs dataset, the changing patterns of curves for both SBFM integrated models and the original ones are nearly the same, showing no notable advantage by SBFM for robustness enhancement. Here we make a note that, when training on Cats-and-Dogs dataset, the strides and pooling coefficients for SBFM are forced to set larger than those on CIFAR-10 and TinyImageNet datasets, due to the memory limitation. This may result in the poor performance of SBFM integrated models on Cats-and-Dogs dataset.

Table~\ref{tab:comparisonFGSM} shows the classification accuracy at $\epsilon$ = 8/255 for both SBFM integrated models and the original ones on three datasets.
From the table, we can see that, on CIFAR-10, VGG16-SBFM is 47.43\% higher accurate than VGG16, achieving 77.44\% classification accuracy. ResNet34-SBFM is 43.46\% higher accurate than ResNet34, achieving 66.41\% classification accuracy. And on TinyImageNet dataset, VGG16-SBFM is 44.37\% higher accurate than VGG16, achieving 72.38\% classification accuracy. ResNet34-SBFM is 71.10\% higher accurate than ResNet34, achieving 81.46\% classification accuracy.
The results demonstrate the binary features learnt by SBFM can enhance the robustness of DCNN models notably.

\begin{table}[ht]
\begin{center}
\begin{threeparttable}

\begin{tabular}{l|c|c|c}
\hline
~ & \textbf{CIFAR-10} & \textbf{TIN} & \textbf{CaD}  \\
\hline
\textbf{VGG16} & 30.01\% & 37.65\% & 7.64\% \\
\textbf{VGG16-SBFM} & \textbf{77.44\%} & \textbf{72.38\%} & \textbf{10.95\%} \\
\hline
\textbf{ResNet34} & 22.95\% & 10.36\% & 1.35\% \\
\textbf{ResNet34-SBFM} & \textbf{66.41\%} & \textbf{81.46\%} & \textbf{2.69\%} \\
\hline
\end{tabular}
\begin{tablenotes}
\item \textbf{TIN} stands for TinyImageNet. \textbf{CaD} stands for Cats-and-Dogs.
\end{tablenotes}
\end{threeparttable}
\end{center}

\caption{Comparison of classification accuracy between SBFM integrated models and original ones under FGSM attack with $\epsilon$ = 8/255.}
\label{tab:comparisonFGSM}
\end{table}

\begin{figure}[h!]
    \centering
    \includegraphics[width=0.25\textwidth]{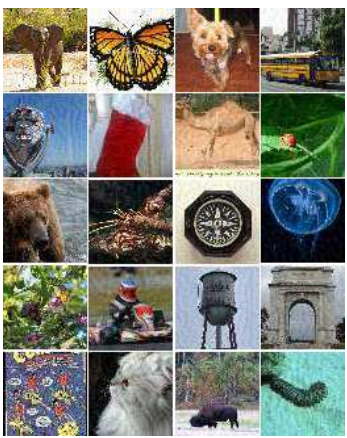}
    \caption{\textbf{Correctly classified adversarial examples of TinyImageNet by VGG16-SBFM}.}
    \label{fig:AEVGGTiny}
\end{figure}

\begin{figure}[h!]
    \centering
    \includegraphics[width=0.25\textwidth]{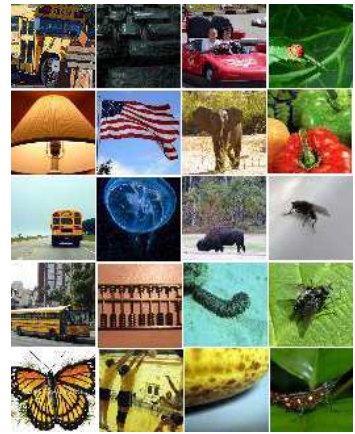}
    \caption{\textbf{Correctly classified adversarial examples of TinyImageNet by ResNet34-SBFM}.}
    \label{fig:AEResTiny}
\end{figure}

\begin{figure}[h!]
    \centering
    \includegraphics[width=0.4\textwidth]{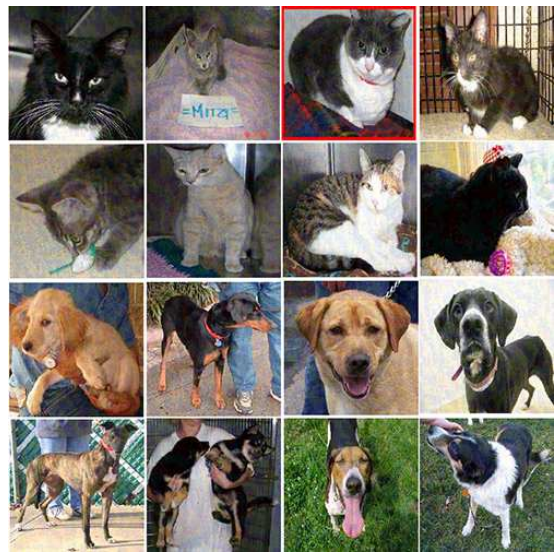}
    \caption{\textbf{Correctly classified adversarial examples of Cats-and-Dogs by VGG16-SBFM}.}
    \label{fig:AEVGGCats}
\end{figure}

\begin{figure}[h!]
    \centering
    \includegraphics[width=0.4\textwidth]{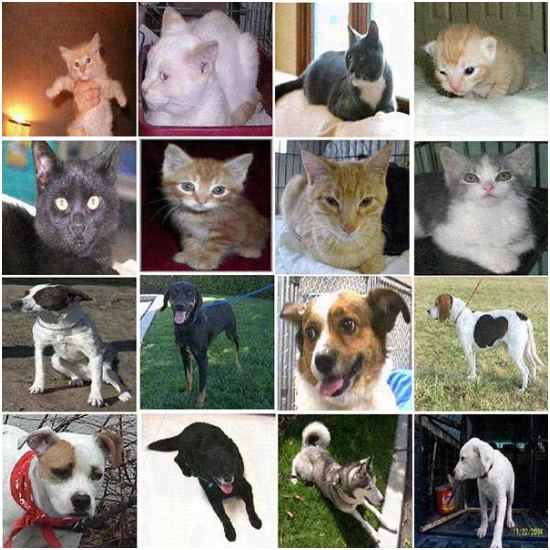}
    \caption{\textbf{Correctly classified adversarial examples of Cats-and-Dogs by ResNet34-SBFM}.}
    \label{fig:AEResCats}
\end{figure}

Fig.~\ref{fig:AEVGGTiny}-Fig.~\ref{fig:AEResCats} show the AEs of $\epsilon$ = 8/255 which can be classified correctly by the SBFM integrated models.

\subsection{Impacts of the number of Sobel layers and threshold}
\begin{figure}[h!]
    \centering
    \includegraphics[width=0.4\textwidth]{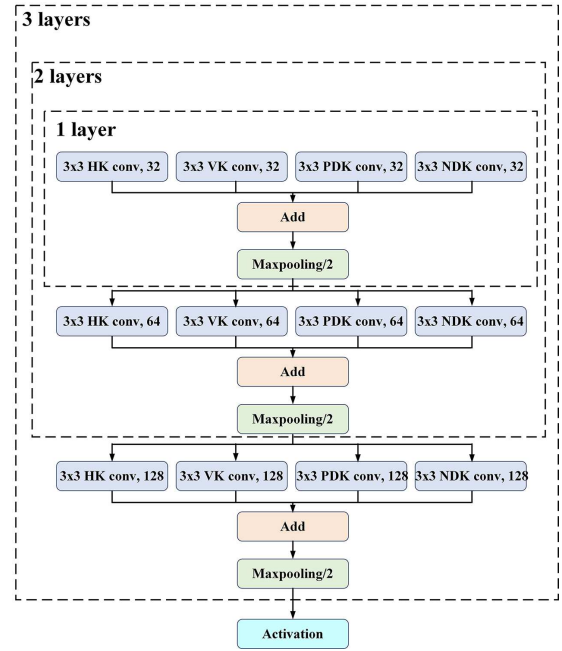}
    \caption{\textbf{SBFM with 1 Sobel layer, 2 Sobel layers, and 3 Sobel layers. HK stands for horizontal kernel. VK stands for vertical kernel. PDK stands for positive diagonal kernel. NDK stands for negative diagonal kernel}.}
    \label{fig:threelayer}
\end{figure}

We examine the impacts of $l$ and $t$ settings on classification accuracy under $\epsilon$=8/255 attack. The values of $l$ and $t$ are set to 1, 2, 3, and 0.4, 0.6, 0.8 respectively.
The loss function, optimizer, the batch size, and the number of epochs are set to be the same.
Each model is run for five times, and the average value is recorded.
Fig.~\ref{fig:threelayer} illustrates the architecture of SBFM with 1 Sobel layer, 2 Sobel layers, and 3 Sobel layers.
Table~\ref{tab:layerthresholdCifar}-Table~\ref{tab:TrainDiffLandT} describe the classification accuracy and training performance of SBFM integrated models under different settings of $l$ and $t$ on CIFAR-10, TinyImageNet, and Cats-and-Dogs datasets.
From the tables, we can see that, the classification accuracy of the SBFM integrated models under attack varies distinctly with different $l$ and $t$ settings, e.g., on CIFAR-10 dataset, ResNet34-SBFM reaches 64.19\% classification accuracy at $l$=2 and $t$=0.8, while achieving 17.19\% classification accuracy at $l$=2 and $t$=0.4.
And in addition, the test accuracy of the SBFM integrated models is not satisfying at certain settings of $l$ and $t$, e.g., in CIFAR-10 dataset, test accuracy of ResNet34-SBFM is 64.53\% at $l$=2 and $t$=0.4, compared to 78.73\% at $l$=2 and $t$=0.8.

\begin{table*}[ht]
\begin{center}
\begin{threeparttable}

\begin{tabular}{l|ccc|ccc|ccc}
\hline
\multirow{2}*{} & \multicolumn{3}{|c|}{\textbf{\emph{l} = 1}} & \multicolumn{3}{|c|}{\textbf{\emph{l} = 2}} & \multicolumn{3}{|c}{\textbf{\emph{l} = 3}}  \\
		\cline{2-10}
		~ & \textbf{\emph{t = 0.4}} & \textbf{\emph{t = 0.6}} & \textbf{\emph{t = 0.8}} & \textbf{\emph{t = 0.4}} & \textbf{\emph{t = 0.6}} & \textbf{\emph{t = 0.8}} & \textbf{\emph{t = 0.4}} & \textbf{\emph{t = 0.6}} & \textbf{\emph{t = 0.8}} \\
\hline
\textbf{VGG16-SBFM} & 40.43\% & \textbf{47.81\%} & 47.22\% & 49.07\% & \textbf{57.36\%} & 52.21\% & \textbf{79.96\%} & 75.08\% & 77.44\% \\
\hline
\textbf{ResNet34-SBFM} & \textbf{62.79\%} & 60.75\% & 58.71\% & 17.19\% & 29.74 & \textbf{64.19\%} & \textbf{68.67\%} & 67.65\% & 66.41\% \\
\hline
\end{tabular}
\end{threeparttable}
\end{center}

\caption{Comparison of classification accuracy of the SBFM integrated models for different $l$ and $t$ settings on CIFAR-10 dataset. $\epsilon$=8/255.}
\label{tab:layerthresholdCifar}
\end{table*}

\begin{table*}[ht]
\begin{center}
\begin{threeparttable}

\begin{tabular}{l|ccc|ccc|ccc}
\hline
\multirow{2}*{} & \multicolumn{3}{|c|}{\textbf{\emph{l} = 1}} & \multicolumn{3}{|c|}{\textbf{\emph{l} = 2}} & \multicolumn{3}{|c}{\textbf{\emph{l} = 3}}  \\
		\cline{2-10}
		~ & \textbf{\emph{t = 0.4}} & \textbf{\emph{t = 0.6}} & \textbf{\emph{t = 0.8}} & \textbf{\emph{t = 0.4}} & \textbf{\emph{t = 0.6}} & \textbf{\emph{t = 0.8}} & \textbf{\emph{t = 0.4}} & \textbf{\emph{t = 0.6}} & \textbf{\emph{t = 0.8}} \\
\hline
\textbf{VGG16-SBFM} & 27.62\% & 31.50\% & \textbf{31.69\%} & 47.28\% & 63.86\% & \textbf{72.38\%} & 35.53\% & 44.68\% & \textbf{50.97\%} \\
\hline
\textbf{ResNet34-SBFM} & 5.21\% & \textbf{25.35\%} & 20.81\% & 40.23\% & 56.68\% & \textbf{65.54\%} & 84.56\% & \textbf{86.86\%} & 81.46\% \\
\hline
\end{tabular}
\end{threeparttable}
\end{center}

\caption{Comparison of classification accuracy of the SBFM integrated models for different $l$ and $t$ settings on TinyImageNet dataset. $\epsilon$=8/255.}
\label{tab:layerthresholdTiny}
\end{table*}

\begin{table*}[ht]
\begin{center}
\begin{threeparttable}

\begin{tabular}{l|ccc|ccc|ccc}
\hline
\multirow{2}*{} & \multicolumn{3}{|c|}{\textbf{\emph{l} = 1}} & \multicolumn{3}{|c|}{\textbf{\emph{l} = 2}} & \multicolumn{3}{|c}{\textbf{\emph{l} = 3}}  \\
		\cline{2-10}
		~ & \textbf{\emph{t = 0.4}} & \textbf{\emph{t = 0.6}} & \textbf{\emph{t = 0.8}} & \textbf{\emph{t = 0.4}} & \textbf{\emph{t = 0.6}} & \textbf{\emph{t = 0.8}} & \textbf{\emph{t = 0.4}} & \textbf{\emph{t = 0.6}} & \textbf{\emph{t = 0.8}} \\
\hline
\textbf{VGG16-SBFM} & 4.83\% & \textbf{18.05\%} & 11.66\% & \textbf{26.31\%} & 24.68\% & 10.95\% & \textbf{45.28\%} & 10.59\% & 7.69\% \\
\hline
\textbf{ResNet34-SBFM} & 1.25\% & 1.28\% & \textbf{1.82\%} & 0.34\% & 1.00\% & \textbf{5.73\%} & 0.05\% & 0.65\% & \textbf{2.69\%} \\
\hline
\end{tabular}
\end{threeparttable}
\end{center}

\caption{Comparison of classification accuracy of the SBFM integrated models for different $l$ and $t$ settings on Cats-and-Dogs dataset. $\epsilon$=8/255.}
\label{tab:layerthresholdCats}
\end{table*}

\begin{table*}[ht]
\begin{center}
\begin{threeparttable}

\begin{tabular}{l|ccc|ccc|ccc}
\hline
\multirow{2}*{} & \multicolumn{3}{|c|}{\textbf{CIFAR-10}} & \multicolumn{3}{|c|}{\textbf{TinyImageNet}} & \multicolumn{3}{|c}{\textbf{Cats-and-Dogs}}  \\
		\cline{2-10}
		~ & \textbf{Tr.Acc.} & \textbf{Te.Acc.} & \textbf{TimePE} & \textbf{Tr.Acc.} & \textbf{Te.Acc.} & \textbf{TimePE} & \textbf{Tr.Acc.} & \textbf{Te.Acc.} & \textbf{TimePE} \\
\hline
\textbf{VGG16-SBFM, $l$=1, $t$=0.4} & 99.17\% & 82.50\% & 23s & 95.96\% & 32.89\% & 81s & 99.58\% & 74.77\% & 22s \\
\textbf{VGG16-SBFM, $l$=1, $t$=0.6} & 99.32\% & 83.07\% & 23s & 95.56\% & 30.63\% & 80s & 96.59\% & 66.80\% & 22s \\
\textbf{VGG16-SBFM, $l$=1, $t$=0.8} & 99.50\% & 83.99\% & 23s & 96.44\% & 30.47\% & 80s & 98.48\% & 75.90\% & 22s \\
\hline
\textbf{VGG16-SBFM, $l$=2, $t$=0.4} & 99.21\% & 83.43\% & 22s & 90.80\% & 31.14\% & 92s & 98.88\% & 71.07\% & 20s \\
\textbf{VGG16-SBFM, $l$=2, $t$=0.6} & 98.87\% & 83.12\% & 22s & 95.42\% & 27.35\% & 93s & 98.90\% & 67.23\% & 20s \\
\textbf{VGG16-SBFM, $l$=2, $t$=0.8} & 99.00\% & 83.24\% & 22s & 92.28\% & 30.01\% & 92s & 99.34\% & 83.17\% & 20s \\
\hline
\textbf{VGG16-SBFM, $l$=3, $t$=0.4} & 99.17\% & 83.48\% & 18s & 56.32\% & 33.00\% & 92s & 99.04\% & 64.27\% & 23s \\
\textbf{VGG16-SBFM, $l$=3, $t$=0.6} & 99.32\% & 83.58\% & 18s & 91.15\% & 30.77\% & 92s & 99.17\% & 82.47\% & 23s \\
\textbf{VGG16-SBFM, $l$=3, $t$=0.8} & 99.19\% & 82.59\% & 18s & 92.03\% & 30.20\% & 92s & 99.64\% & 82.80\% & 23s \\
\hline
\hline
\textbf{ResNet34-SBFM, $l$=1, $t$=0.4} & 99.59\% & 72.80\% & 48s & 98.54\% & 15.77\% & 247s & 97.89\% & 62.07\% & 86s \\
\textbf{ResNet34-SBFM, $l$=1, $t$=0.6} & 99.43\% & 77.92\% & 48s & 98.33\% & 27.81\% & 246s & 98.12\% & 60.77\% & 87s \\
\textbf{ResNet34-SBFM, $l$=1, $t$=0.8} & 99.32\% & 78.96\% & 48s & 98.31\% & 24.93\% & 246s & 98.04\% & 56.43\% & 87s \\
\hline
\textbf{ResNet34-SBFM, $l$=2, $t$=0.4} & 99.50\% & 64.53\% & 50s & 98.22\% & 22.36\% & 250s & 98.69\% & 71.07\% & 89s \\
\textbf{ResNet34-SBFM, $l$=2, $t$=0.6} & 99.34\% & 65.87\% & 50s & 97.98\% & 24.20\% & 250s & 98.12\% & 68.17\% & 89s \\
\textbf{ResNet34-SBFM, $l$=2, $t$=0.8} & 99.67\% & 78.73\% & 50s & 97.87\% & 21.73\% & 249s & 97.99\% & 65.87\% & 89s \\
\hline
\textbf{ResNet34-SBFM, $l$=3, $t$=0.4} & 99.43\% & 69.49\% & 51s & 97.65\% & 21.78\% & 254s & 96.03\% & 75.10\% & 91s \\
\textbf{ResNet34-SBFM, $l$=3, $t$=0.6} & 99.17\% & 73.40\% & 51s & 96.92\% & 23.68\% & 254s & 96.64\% & 71.33\% & 90s \\
\textbf{ResNet34-SBFM, $l$=3, $t$=0.8} & 99.27\% & 74.58\% & 51s & 97.92\% & 26.31\% & 256s & 97.13\% & 73.80\% & 91s \\
\hline
\end{tabular}
\begin{tablenotes}
\item \textbf{Tr.Acc.} stands for training accuracy. \textbf{Te.Acc.} stands for test accuracy. \textbf{TimePE} stands for training time per epoch.
\end{tablenotes}

\end{threeparttable}
\end{center}

\caption{Comparison of training performance between SBFM integrated models with different $l$ and $t$ settings.}
\label{tab:TrainDiffLandT}
\end{table*}

\section{Discussions}
The experimental results from Section~\ref{sec:Expe} shows SBFM has no side effect on model training, and it is lightweight and can be integrated into any backbone. Furthermore, the binary features learnt by the SBFM can enhance the robustness of DCNN models notably.

Here, we want to emphasize that, in this paper, a heuristic combination form of texture features and binary features (shown in Fig.~\ref{fig:Archi}) is introduced. We believe it is worthwhile to explore other effective and efficient combination forms of both features to enhance the robustness of DCNN models. Also, from the experimental results, it can be seen that, the performance of SBFM integrated models under attack changes distinctly for different parameter settings, and the generalization ability of the trained SBFM integrated models is relatively weak at certain parameter settings.
It hints us that it is of great importance to design an optimization framework for parameter searching to yield the SBFM integrated models with exceptional classification accuracy under attack and satisfying generalization ability.

\section{Conclusions}
Enhancing the robustness of DCNN models is of great significance for the safety-critical applications in the real world. In this paper, we first raise the question "can a DCNN model learn certain features which are insensitive to small perturbations, and further defend itself no matter what attack methods are presented", and make a beginning effort to answer it. A shallow binary feature module is proposed, which is lightweight and can be integrated into any popular backbone. The experimental results on multiple datasets show the binary features learnt by the shallow binary feature module can enhance the robustness of DCNN models notably. In future's work, we endeavor to explore other effective and efficient combination forms of binary features and texture features, and design an optimization framework for the parameter searching to yield models with good performance under attack and generalization ability. The work in this paper shows it is promising to enhance the robustness of DCNN models through feature learning.

\bibliographystyle{IEEEtran}
\bibliography{egbib}

%

\end{document}